\documentclass[a4paper,twoside,10pt]{article}
\usepackage[T1]{fontenc}
\usepackage[latin1]{inputenc}
\usepackage{amsmath, amssymb, latexsym, amsthm, mathrsfs}
\usepackage{graphicx}
\usepackage{ulem}
\usepackage{dsfont}
\input{epsf}
\usepackage[lined]{algorithm2e}
\usepackage{hyperref}

\date{}
\def\XXint#1#2#3{{\setbox0=\hbox{$#1{#2#3}{\int}$}
\vcenter{\hbox{$#2#3$}}\kern-.5\wd0}}

\renewcommand{\epsilon}{\varepsilon}
\renewcommand{\phi}{\varphi}

\DeclareMathOperator*{\argmin}{argmin}

\theoremstyle{plain}

\theoremstyle{remark}

\usepackage{listings}
\usepackage{color}
\title{Partial Wasserstein and Maximum Mean Discrepancy distances for bridging the gap between outlier detection and drift detection}

\author{Thomas Viehmann
\thanks{MathInf GmbH, tv@mathinf.eu}
}
\date{June 2021}
\begin{document}

\maketitle

\begin{abstract}
  With the rise of machine learning and deep learning based applications in practice, monitoring, i.e. verifying that these operate within specification, has become an important practical problem.
  An important aspect of this monitoring is to check whether the inputs (or intermediates) have strayed from the distribution they were validated for, which can void the performance assurances obtained during testing.
  
  There are two common approaches for this. The, perhaps, more classical one is outlier detection or novelty detection, where, for a single input we ask whether it is an outlier, i.e. exceedingly unlikely to have originated from a reference distribution. The second, perhaps more recent approach, is to consider a larger number of inputs and compare its distribution to a reference distribution (e.g. sampled during testing). This is done under the label \textit{drift detection}.

  In this work, we bridge the gap between outlier detection and drift detection through comparing a given number of inputs to an automatically chosen part of the reference distribution.
\end{abstract}

When we deploy programs based on machine-learning models in the real world, a crucial question we have is if the data and model operations are within spec, i.e. if the observed data and model behviour matches those during model training and validation. Following the work of \cite{FailingLoudly}, one common method of detecting deviations is to apply two-sample tests to reference and observed data or intermediate or output valuees of the model. But while making use of two-sample tests is indeed a very promising approach, so far, the approach has been insufficiently adapted to the actual problem at hand, leading to insufficient performance in practical applications.

Our contribution in this paper is that we modify the two-sample testing approach to make it more fitting to the actual question asked in drift detection. The key steps we contribute are:

\begin{itemize}
\item We use the partial Wasserstein distance as a test statistic for the two-sample testing matching part of the reference distribution.
  This bridges two-sample-test-based drift detection and outlier detection.
\item Inspired by the partial Wasserstein distance, we introduce a partial matching for the MMD distance. As the most natural definition involves a computationally very expensive quadratic programming problem, we use optimal transport problem underlying the partial Wasserstein distance to give an approximation.
\item We emphasize the role of the two samples in the testing process as coming from the reference distribution and the testing distribution to refine the bootrapping procedure to generate $p$-values.
\end{itemize}

\section{The need to bridge outlier detection and drift detection}

We briefly describe two properties of practical applications statistical monitoring of model inputs that indicate that two-sample testing as described by \cite{FailingLoudly} does not fully meet the requirements for detecting drift. One is an à priori consideration, while the other is an empirical observation from practical use of the method.

From a philosophical perspective, we see the following deficit in framing drift detection as a two-sample problem. The null hypothesis we statistically test against is that the reference and test samples have  been drawn from the same distribution. However, there are many reasons why we believe that this assumption is often deeply unrealistic:

\begin{itemize}
\item There may be fluctuations in the environment that mean that the reference dataset is more rich than we expect the test dataset to be. For example, outdoor lighting conditions vary over time through day and night, with weather conditions, or the seasons of the year. If our model is sufficiently trained on to cover all of these, it will operate in normal conditions even though the test samples, drawn from a much shorter time interval than the reference, do not show this variety. Now, we could make the variation explicit and then test against conditional references, but this would cause significant additional effort.
\item Inputs provided by human users, e.g. search queries, are likely to have spikes for user interests in a given time range rather than uniformly querying the entire reference
  (e.g. the contents of a knowledge database). This, too, should count as normal operation. 
\end{itemize}

One way to mitigate these effects could be to enlarge the data sample (and the time to collect it), but this may render timely drift detection infeasible.

Note how outlier detection would, in the situations described above, conclude that the inputs are not outliers. However, we still want the model monitoring to consider a collapse of the input distribution (e.g. a frozen camera image from a visual inspection system) to be flagged as problematic, but an outlier detection, unaware of the distribution of the inputs beyond a single given sample, cannot identify this. In this sense, we need a bridge between drift detection and outlier detection: We want the multi-sample approach on the test side like drift detection. At the same time we aim to remove the requirement to compare against the full test distribution, a requirement that outlier detection does not have. 

We believe that this à priori consideration highlights at least partially the cause of the following observation from applying drift detection to practical problems:
The statistical testing often leads to the rejection of the hypothesis that the reference and test distribution are the same with extremely high confidence (extremely low p-values like $10^{-10}$) even when human inspection would conclude that no relevant drift has taken place. This appears to be an indication that the hypothesis we are testing is not well-matched to what we actually want to detect.
In practice, we may \textit{calibrate} the drift detection to consider values of the test statistic observed during normal operation, say at the beginning or during the test of a deployment, but this may lead our testing to miss actual cases and is not very satisfying philosophically.

With this in mind, we introduce the mathematical tool that will give us the bridge we seek.

\section{Wasserstein distance and two-sample testing}

Of the test setups discussed in \cite{FailingLoudly}, simultaneous Kolmogorov-Smirnov-Tests on the marginals and Kernel maximum mean discrepancy (MMD) tests are very popular and are implemented by libraries (e.g. Alibi Detect \cite{alibidetect} and TorchDrift \cite{torchdrift}). To prepare for bridging drift and outlier detection, we use the Wasserstein distance as test statistic instead.

The $p$-Wasserstein distance between two probability measures $\mu, \nu$ with finite $p$-th moments on a metric space $M$ with distance $d$ is defined as
\[
  \mathcal{W}_p(\mu, \nu) = \inf_{P \in U(\mu, \nu)} \left( \int_{M \times M} d(x, y)^p dP(x,y) \right)^{1/p},
\]
where the set of couplings $U(\mu, \nu)$ is given as set of the probability measures on $M \times M$ with marginals $\mu$ and $\nu$ when integrating
over the second and first factors, respectively, i.e.
\[
  \int_{M} P(x,y) dy = \mu(x) \text{ and } \int_{M} P(x,y) dx = \nu(y).
\]
Here, the equality of the measures is to be understood in the sense of distributions.

An entropy-regularized version that (in the discrete case) computationally attractive has been championed by \cite{Cuturi}, for a comprehensive overview see \cite{PeyreCuturi} and has seen good success in various applications. The regularization has various desirable properties (better computational efficiency and also sample efficiency for estimation, and, with a slight modification to Sinkhorn Divergences).
It also provides a bridge between the Wasserstein distance and the Maximum Mean Discrepancy distance and by taking the limit (to $0$ and $\infty$ respectively, see \cite{Genevay}). Unfortunately, it turns out that
for our purposes, the na\"ive Sinkhorn algorithm does not converge well, apparently due to the near-singularity of the our transport problem. We have not attempted whether more advanced versions, such as the subspace Robust Wasserstein distance \cite{SRW} offer a solution to the convergence difficulties.

The use of the (regularized) Wasserstein-distance for two-sample testing is discussed in \cite{Ramdas}. A disadvantage of the (unregularized or weakly regularized) Wasserstein distance compared to the MMD is is the sample complexity, i.e. how fast or slow the sampled distance converges to the distance between the underlying distributions. This convergence rate for the Wasserstein Distance is $O(n^{-\frac{1}{d}})$, with $n$ being the smaple size and $d$ the dimension, compared to $O(n^{-\frac{1}{2}})$ for MMD distances with suitable kernels (see \cite{PeyreCuturi} for a precise statements and a review of the literature). The regularized Wasserstein distance exibits intermediate sample complexity  \cite{Genevay}.

One aspect where the practical implementation of a test based on the Wasserstein distance differs from that of MMD is in how the details of the methods are chosen.
To operationalize the Wasserstein distance, one needs to choose the exponent $p$ and, if desired, the regularization parameter $\varepsilon$. For the MMD a choice of kernel is required. Many commonly used kernels (e.g. the squared exponential (Gaussian) or the exponential (Laplace) kernel) have a lengthscale parameter that needs to be selected and is expected to be rather data-specific.
This is not completely set in stone - e.g. we could use MMD with the kernel derived from the regularized Wasserstein distance in the limiting process $\varepsilon \rightarrow \infty$.

There is another key difference: For the Wasserstein distance, the coupling matches points from the two distribution, in the classical Monge formulation we even seek a transport map form one to the other.
This allows us to decompose the Wasserstein (or transportation cost) distance into points in one distribution that causes it. The spreading out of the entropy regularization, however, spreads the mapping across the mass. In the limit $\varepsilon \rightarrow \infty$, i.e. the limit towards MMD, the coupling is uniform, so that this decomposition is not as valuable. This ability to use the decomposition is crucial for our partial matching.

\section{Partial matching of the reference distribution}
\label{sec:partial_wasserstein}

If we were to create a distance-based outlier detector, we could use the distance of the single-item test sample to the closest reference sample as a criterion. We can generalize this to a larger test samples by trying to match them such that the total distance is small. 

Partial Wasserstein distance and unbalanced optimal transport has been used in the literature (e.g. \cite{CaffarelliMcCann}, \cite{Chapel}), we briefly recount the definition and motivation here.
Given our observation in the last section that we can use the coupling to extract the distance contribution attributable to given points, we now just need some ``default match'' for the unmatched proportion of the reference distribution. As we want true true test points to match to their closest counterparts in the reference without interferance from these default matches, an \textit{infinitely far away} mass would suit our needs. In practice, we approximate this by putting the the probability mass to not be matched on a very distant point. For the unregularized Wasserstein distance, making this point more distant than the maxmimal distance between reference and test sample would be sufficient. (With regularization, we can expect some distortion of the distance contributions because some reference probability mass close to the test samples will be matched with the distant point instead to increase entropy. On the other hand, we may expect the Sinkhorn algorithm to converge more slowly when the distance to the far away point increases the condition of the cost matrix, and as mentioned above, our experiments with the plain regularized algorithm showed that the convergence was prohibitively slow.)

As we wish to work with the Wasserstein distance in the two sample setting, we us take the reference sample $X_j$, $j=1,...,N$ and define a uniform distribution $\mu = \frac{1}{N} \sum_{i = 1}^N \delta_{X_i}$, with $\delta_x$ the dirac distribution on the point $x$. On the test sample $Y_{j}$, $j=1,...,M$ we put uniform weights summing to $\alpha$. The distant point $Z$ gets assigned the remaining $1-\alpha$ probability mass, so we have the test distribution $\nu = \frac{\alpha}{M} \sum_{j=1}^M \delta_{Y_j} + (1-\alpha) \delta_{Z}$. The algorithm's input are the the weights and the transportation cost between the points, which we set to
\[
  C_{ij} =
  \begin{cases}
    |X_i - Y_j|^p & \text{ for }i = 1,...,N \text{ and } j=1,...,M, \\
    D^p & \text{ for } i = 1,...,N \text{ and } j = M+1.
  \end{cases}
\]
We choose the distance of the far away point to be $D = 1.1 \max_{ij} |X_i - Y_j|$, other choices are possible.
We find the minimizing coupling $P_{ij}$ and then sum the associated cost except the column of the distant point and take the $1/p$-th power, so the distance between the samples is
\[
(  W^\alpha_p (X, Y))^p = \frac{1}{\alpha} \sum_{i=1}^N \sum_{j=1}^M C_{ij} P_{ij}.
\]

Heuristically, we may expect the partial matching property to mitigate the sample complexity issue with only having a relatively small number of sample points in the test distribution: While few test sample points cannot be expected to cover the target distribution well, only the partial match to close-by parts of the reference distribution are needed.

This bridges drift detection and outlier detection in the following sense: When the $\alpha = 1$, we have the classical two-sample tests of drift detections. If, for $M = 1$ and given $N$, we set $\alpha = \frac{1}{N}$, the result will be the distance to the closest point $X_i$. This can be seen as a quantity for outlier detection, either directly or by interpreting it as an approximation to a logarithmic kernel-density estimate at $Y_1$ (implicitly assuming that there is only a single point $X_i$ close to $Y_1$), varying the value of $\alpha$ betwee nthese two extremes provides the bridge.

\section{Partial MMD matching}
\label{sec:partial_mmd}

In our experimentation, reported in Section \ref{sec:experiment}, using the MMD distance appears to lead to detectors that can more reliably detect drift than
those built using the Wasserstein distance. However, the distant point trick to match against only a relevant part of the reference does not work with the MMD distance. In this section, we formulate the partial matching as an optimization problem instead: We add weights to the reference points in the kernel-based MMD formula. Optimizing these weights leads to a constrained quadratic programming problem. Because solving the quadratic programming is computationally very expensive, we use the optimal transport problem of the last section: We use the coupling from the partial Wasserstein distance as a set of weights and obtain an upper bound for (and heuristic approximation of) the partial MMD distance.

\subsection{Partial MMD distance through quadratic programming}

Recall that the (biased) empirical MMD distance from \cite{Gretton} is
\[
MMD^2(X, Y) = \frac{1}{n^2} \sum_{i} \sum_{j} k(x_i, x_j)
           + \frac{1}{m^2} \sum_{i} \sum_{j} k(y_i, y_j)
           - 2 \frac{1}{n m} \sum_{i} \sum_{j} k(y_i, x_j),
\]
where $k$ is the Kernel. Writing the kernel evaluations using
matrices $K^X_{ij} = k(x_i, x_j)$, $K^Y_{ij} = k(y_i, y_j)$ and $K^{YX}_{ij} = k(y_i, x_j)$ and weight vectors
$w = \left(\begin{smallmatrix}1/n\\...\\1/n\end{smallmatrix}\right) \in \mathbb{R}^n$
$v = \left(\begin{smallmatrix}1/m\\...\\1/m\end{smallmatrix}\right) \in \mathbb{R}^m$, we can rewrite the this as
\begin{equation}
  \label{eq:mmd_matrix}
MMD^2(X, Y) = w^T K^X w + v^T K^{Y} v - 2 v^T K^{YX} w.
\end{equation}

To allow matching on a subset of $X$, we optimize over $w$ with suitable conditions instead of fixing it. For the fraction $0 < \alpha \leq 1$  we obtain the $\alpha$-partial MMD distance
\begin{equation}
\label{eq:qp}
(MMD^{\alpha}(X, Y))^2 = \min_w  w^T K^X w + v^T K^{Y} v
        -  2 v^T K^{YX} w,
\end{equation}
where $w$ is minimized subject to
\begin{equation}
\label{eq:constraints}
 0 \leq w_i \leq \frac{1}{\alpha n}, \qquad \sum_i w_i = 1,
\end{equation}
with $v = \left(\begin{smallmatrix}1/m\\...\\1/m\end{smallmatrix}\right)$.
Note that by design, if $\alpha$ is $1$, the constraints impose $w_i = \frac{1}{n}$, while for $\alpha < 1$ we may overweight some components at the expense of others.

As the kernel Matrix $K^X$ is symmetric positive definite,
we may take the Cholesky decomposition $K^X = R^T R$ and then reduce the quadratic programming problem into an equivalent least squares problem (with a different constant offset)
\[
w = \argmin_w \| R w - R^{-T} (v^TK^{YX})   \|^2 
\]
subject to the same constraints.

With the generic quadratic programming or constrained least squares solvers in common software packages like SciPy, using this approach appears to be prohibitively costly compared to the two-stage method, so we did not persue it further (in the base case a single evaluation of the $MMD^\alpha$ took about 2.5 seconds, this compares to less than 0.14 seconds for the four methods in our empirical evaluation. While this direct approach is theoretically more satisfying than the two-stage approach, this implementation concern led us to not consider it further.

\subsection{Ad hoc optimization}

We may formulate an ad hoc optimization taking Newton-type steps on the coordinates which are not on the boundary. One of the choices our in our implementation is to use a fixed (heuristic) size for the loops in order to minimize
GPU synchronizations.

We take as set of \textit{working coordinates} the coordinates which are not on the boundary of the admissible set and a (pseudo-randomized) subset of coordinates on the boundary for which the gradients are strongly pointing inwards.
Then we compute an update direction with the Newton update on the working coordinates.
To deal with distortions from (approximately) enforcing the constraints \eqref{eq:constraints}, we reduce step sizes and see if we actually get an improvement.
We empirially saw that Newton steps can get stuck, so we also do a number of gradient steps.

Our exact procedure is given in Algorithm \ref{alg:adhoc-opt}.

\begin{algorithm}
  \SetAlgoLined
  \SetKwComment{comment}{\# }{}
  \KwIn{Kernel matrices $K^X$, $K^Y$, $K^{XY}$, fraction $\alpha$ to match}
  \KwOut{Approximate minimizing weight $w$}
  $w$ := $(1/n,...1/n) \in \mathbb{R}^n$\;
  $\widehat{MMD^\alpha}^2(w)$ := $w^T K^X w + v \sum K^Y v^T - 2 w^T K^{XY} v$\;
  \For{$N_{iter}$ iterations}{
       $r$ := random number $0...1$\;
       $gr$ := $2 K^X w - 2 K^XY v$\;
       $gr_{min}$ := $\min \{gr_i | w_i < \frac{1}{\alpha n}\}$\;
       $gr_{max}$ := $\max \{gr_i | w_i > 0\}$\;

       $working$ := $\{i | ((w_i > 0) \text{ or } (gr_i < gr_{min} * r))  \text{ and }  ((w_i < \frac{1}{\alpha n})  \text{ or } (gr_i > gr_{max} * r))$\;

       $upd_{working}$ := $(K^X_{working})^{-1} gr_{working}$; \comment{implemented using Cholesky }

       \For{$step \in \{ 1, 1/5, 1/25, ..., 1/5^4 \}$}{
         $w_{cand}$ = $w - step * upd_{working}$\;
         $w_{cand}$ = clamp($w_{cand}$, min=$0$, max=$\frac{1}{\alpha n}$)\;
         $w_{cand}$ = $w_{cand} / \sum w_{cand}$; \comment{this potentially lets $w$ violate the upper bound}
         \If{$\widehat{MMD^\alpha}^2(w_{cand}) < \widehat{MMD^\alpha}^2(w)$ }{
           $w$ := $w_{cand}$\;}
       }

       recompute $gr$, $gr_{min}$, $gr_{max}$, and $working$ \;
       $upd_{working}$ := $gr_{working} - \frac{1}{\#working} \sum gr_{working}$\;
       \For{$step \in \{ 0.1, 0.1/5, 0.1/25, ..., 0.1/5^4 \}$}{
         $w_{cand}$ = $w - step * upd_{working}$\;
         $w_{cand}$ = clamp($w_{cand}$, min=$0$, max=$\frac{1}{\alpha n}$)\;
         $w_{cand}$ = $w_{cand} / \sum w_{cand}$\;
         \If{$\widehat{MMD^\alpha}^2(w_{cand}) < \widehat{MMD^\alpha}^2(w)$ }{
           $w$ := $w_{cand}$\;}
       }
     }
     \KwRet{$w$}

  \caption{Ad-hoc constrained optimization using PyTorch CUDA primitives}\label{alg:adhoc-opt}
\end{algorithm}

In our empirical observation, this, delivered a good approximation to the results computed by stock quadratic programming as described above, but is also considerably slower than the optimal transport based upper bound below.

\subsection{Optimal transport based upper bound for the partial MMD}

To avoid the computational burden of the quadratic programming problem,
we use the much cheaper partial Wasserstein distance minimization problem
to obtain a feasible vector $w$ in the MMD minimization problem and thus
an upper bound to the partial MMD distance $MMD^\alpha$.

Recall that Wasserstein coupling $P_{ij}$ sums to the marginals.
In particular, $w_i = \frac{1}{\alpha} \sum_{j=1}^M P_{ij}$ is admissible in the quadratic programming problem \eqref{eq:qp}, i.e. statisfies the constraints \eqref{eq:constraints}, giving an upper bound to the minimum.
Plugging this $w$ into the MMD formula \eqref{eq:mmd_matrix} gives us the approximate partial MMD drift detector.

The MMD distance bounds obtained through this method do not appear to be a good approximation to the partial MMD distances, but they do result in a test that performed reliably in our experimental evaluation below.

\section{$p$-values: Revisiting bootstrapping}
\label{sec:bootstrapping}

In the author's practical experience, people feel more comfortable when discussing $p$-values than using the distances or test statistics directly. As the rate of false positives, the $p$-values that are of practical use in this exercise tend to be much lower than those used in other situations because here the testing is repeated rather often (i.e. a monitoring tool testing for drift several hundreds of times a day would not work well with, say, a threshold corresponding to $1\%$).

The modular view on getting a $p$-value, also proposed by \cite{FailingLoudly} for the kernel MMD test, is to reduce the drift detection to two-sample testing. This two-sample testing is used as a black box and thus the two-sample test null hypothesis \textit{the reference and the test data have originated from the same distribution} is used. Bootstrapping is then done by randomly swapping $X_i$ and $Y_i$, implemented through drawing random permutations of the joint sequence $X_1, ..., X_N, Y_1, ..., Y_M$ and then taking the first $N$ and last $M$ to compute the test statistic.

We can, however, reformulate the null hypothesis to be slightly closer to the original drift detection task: \textit{the test data has originated from the reference distribution}.
This suggests that instead of waiting for test time, we draw $N+M$ samples from the reference distribution and bootstrap the test statistic's distribution from these.
If the null hypothesis were true, this would be exactly equivalent, however, when the test data is from a different distribution, the ``modular'' approach leads to the null hypothesis that both come from a ``mix'' of reference and test distribution.

As the computationally expensive permutation testing has to only be done once, this also allows for efficiency gains if many drift tests are to be made, which we expect to be the case in model monitoring.
The additional cost for this approach, the need to an provide additional $M$ samples from the reference distribution, apprears to be affordable in practical applications. We can combine this with fitting an appropriate distribution to the bootstrapped empirical distribution. For MMD, we have seen good results with a shifted gamma distribution (following \cite{Gretton}, but adding the shift for the expected minimal value of the test statistic as it seemed to improve the fit).
This fit also seems broadly reasonable for the (partial) Wasserstein distance, so we adopt it there, too. This introduces an artificial cut-off point on the left hand side and, indeed, in particular for the partial matching, a normal approximation may be better. We are, however, more interested in the right hand side approximation and for the
full Wasserstein distance, we observed that the shifted gamma distribution provides a better fit. Figure \ref{fig:goodness-of-fit} shows representative fits from our experiments.

Our empirical evaluation suggests that the additional sharpness of the test of the modified bootstrap improves the makes drift detection more reliable.

\section{Empirical evaluation}
\label{sec:experiment}

We evaluate our approach on the STL10 dataset \cite{STL10}. This dataset was chosen to balance the three following criteria:

\begin{itemize}
\item The dataset is readily accessible. We use the operating description of \textit{included in the TorchVision library} here.
\item We wanted to have largeish pictures. While STL10 image size of 96x96 is still smaller than ImageNet's 224x224 standard size, we consider it an improvement over the MNIST and CIFAR10 datasets used by \cite{FailingLoudly}.
As practicioners, we observed that some techniques (notably the \textit{untrained autoencoders} of \cite{FailingLoudly}) appear to work less well on real-world (largeish) images than experiments on MNIST and CIFAR10 suggest. To re-use a standard pretrained model, we scale the images to 224x224 before feeding them into the network. The artificial drift is applied after scaling.
\item We wanted to use a reasonably small dataset, we expect our results to not depend on arbitrarily large datasets and want our work to be accesible to resource-constrained members of the target audience.
\end{itemize}

Our comparison point is the kernel MMD distance as implemented by \texttt{torchdrift} \cite{torchdrift}. This largely follows the implementation provided with \cite{FailingLoudly}, but has a lengthscale heuristic closer to the original MMD code \cite{Gretton}.
We base our model on the pre-trained (on ImageNet) ResNet 18 backbone from TorchVision. Following the fine-tuning approach, we re-use the convolutional part as is and train a two-layer head, with intermediate activation size $32$.
We use this fine-tuned network up to this $32$-dimensional activation as our feature extractor. The reference distribution is taken as 1000 random images from the training set. All testing is done against samples from the test set. We run the test on 1000 balanced and 1000 imbalanced sample draws. For the balanced sample draws, we draw randomly from the test distribution. For easy implementation with PyTorch dataloaders, we sample with repetition. For the imbalanced samples, we stratify the test set into the 10 classes and for each of them draw 100 batches from that said class (again, with repetition). These samples are then corrupted by gaussian blur (of a severity level matched to parameters modelled after \cite{Hendrycks}, implemented in TorchDrift). Over the 10 trial runs, the trained models achieve a mean accuracy of $93.6\%$ on the uncorrupted test set. This drops to $85.6\%$ with the baseline blur (strength 2) and $91.1\%$ and $74.7\%$ with the weaker stronger blur sensitivities (strength 1 and 3, respectively).

We evaluate four drift detectors:
\begin{itemize}
\item The MMD drift detector, as presented in \cite{FailingLoudly} and implemented in TorchDrift,
\item a drift detector using the Wasserstein metric (matching the full reference distribution),
\item the drift detector of Section \ref{sec:partial_wasserstein} using the Wasserstein metric matching a part of the reference distribution,
\item the partial MMD drift detector of Section \ref{sec:partial_mmd} in three variants, the \textit{two-stage} setup using the Wasserstein coupling to compute the weight,
\item the \textit{approximative} partial MMD using the ad-hoc constrained optimization, and
\item the partial MMD computed using a \textit{QP}-package (\texttt{qpsolver}) to solve the constrained least squares problem. As this variant takes much more time than the others,
  we only computed this for the base scenario (not the sensitivities) and only computed every fifth test batch to derive the statistics presented below. Also we used only 100 bootstrap permutations to compute the test statistic under the null hypothesis.
\end{itemize}

\begin{figure}
\includegraphics[width=5cm]{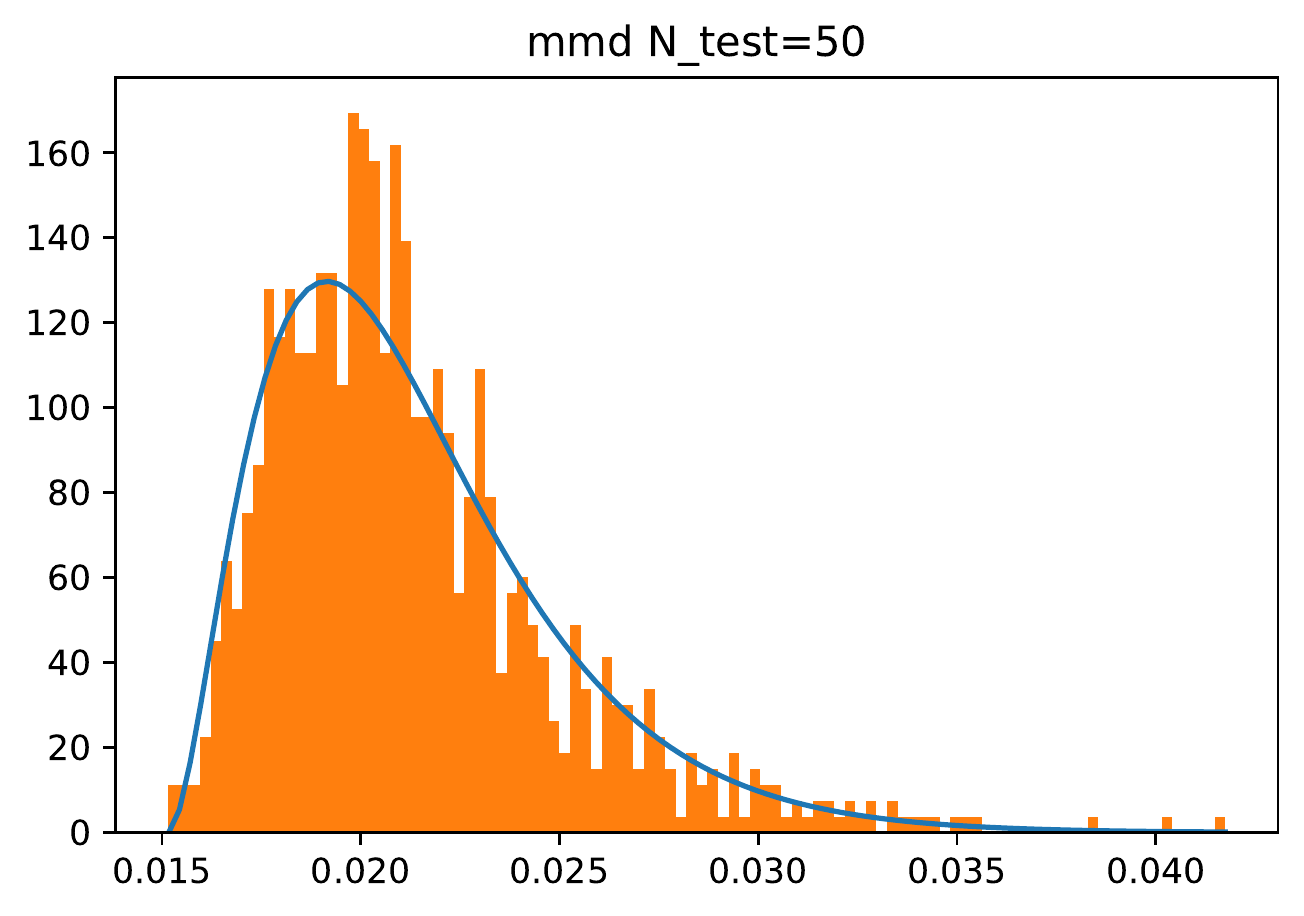}
\includegraphics[width=5cm]{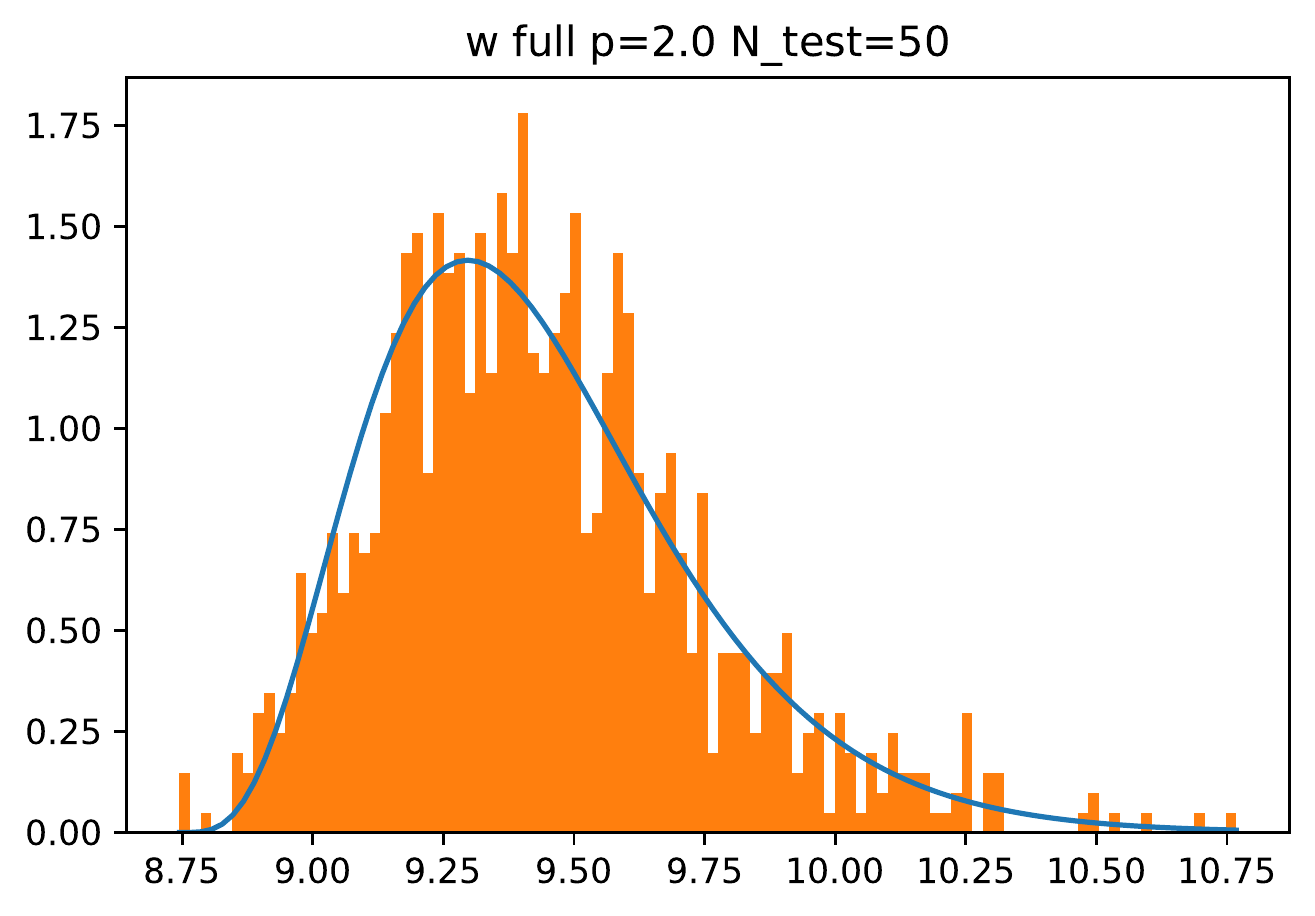}
\includegraphics[width=5cm]{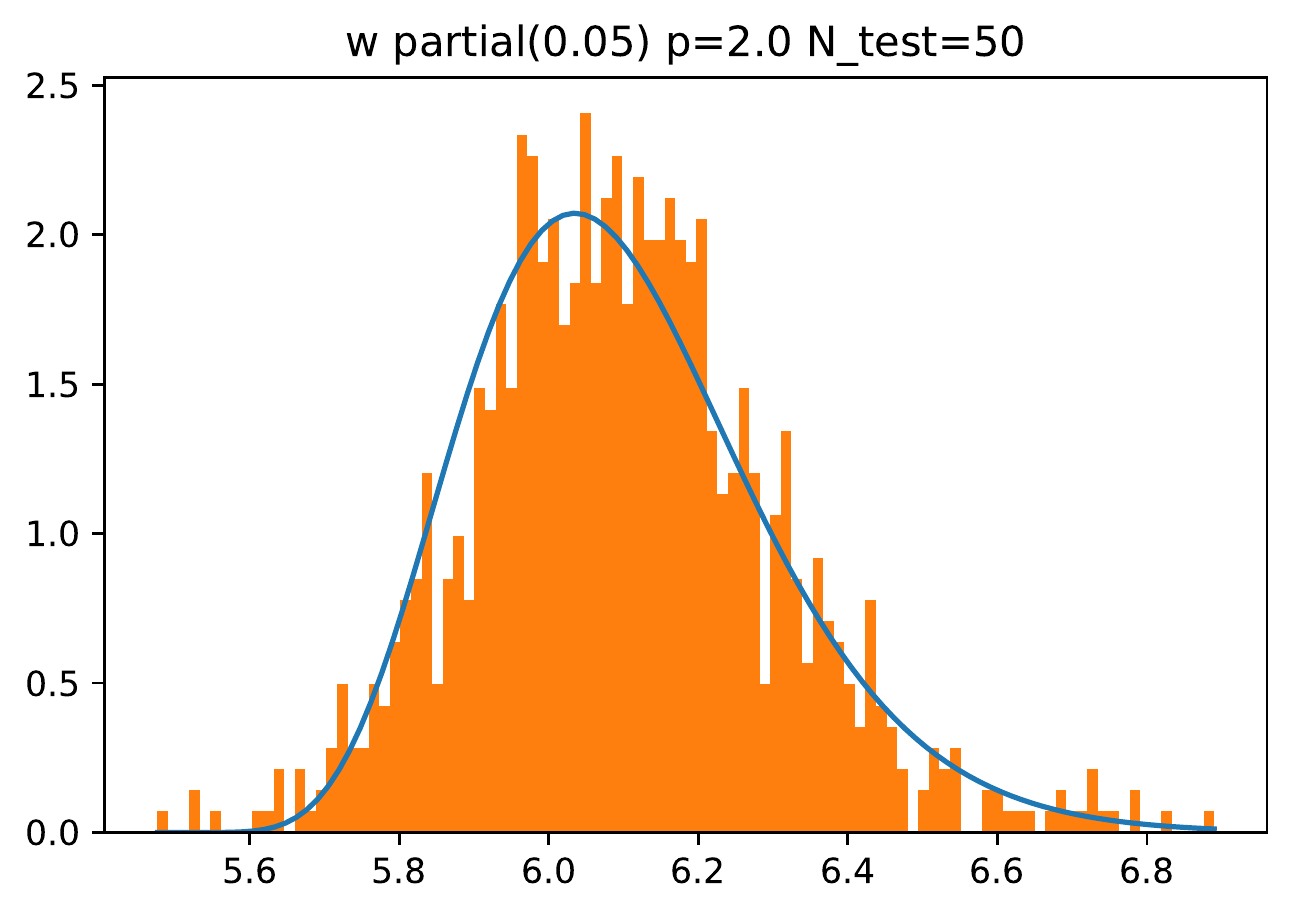} \\
\includegraphics[width=5cm]{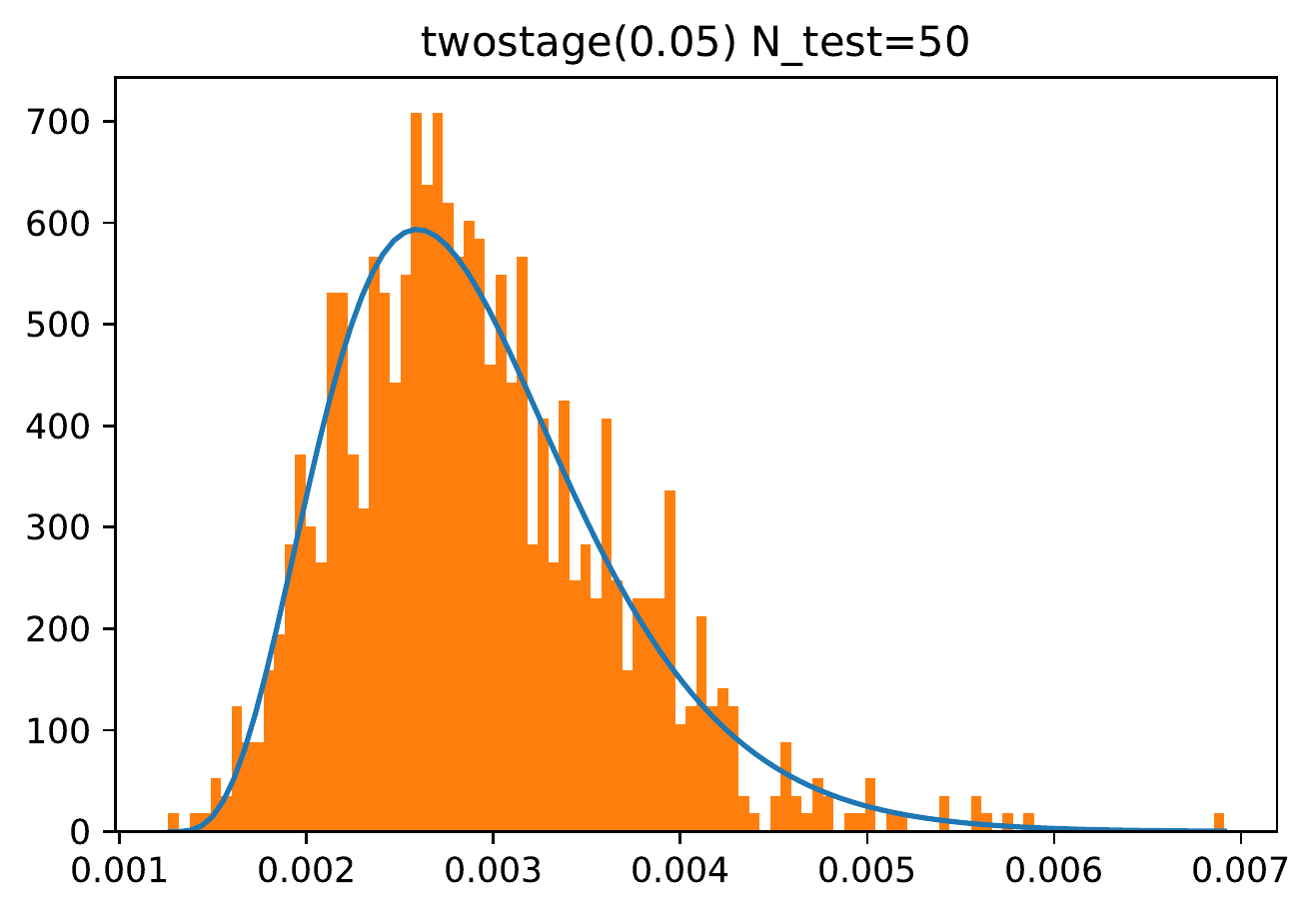}
\includegraphics[width=5cm]{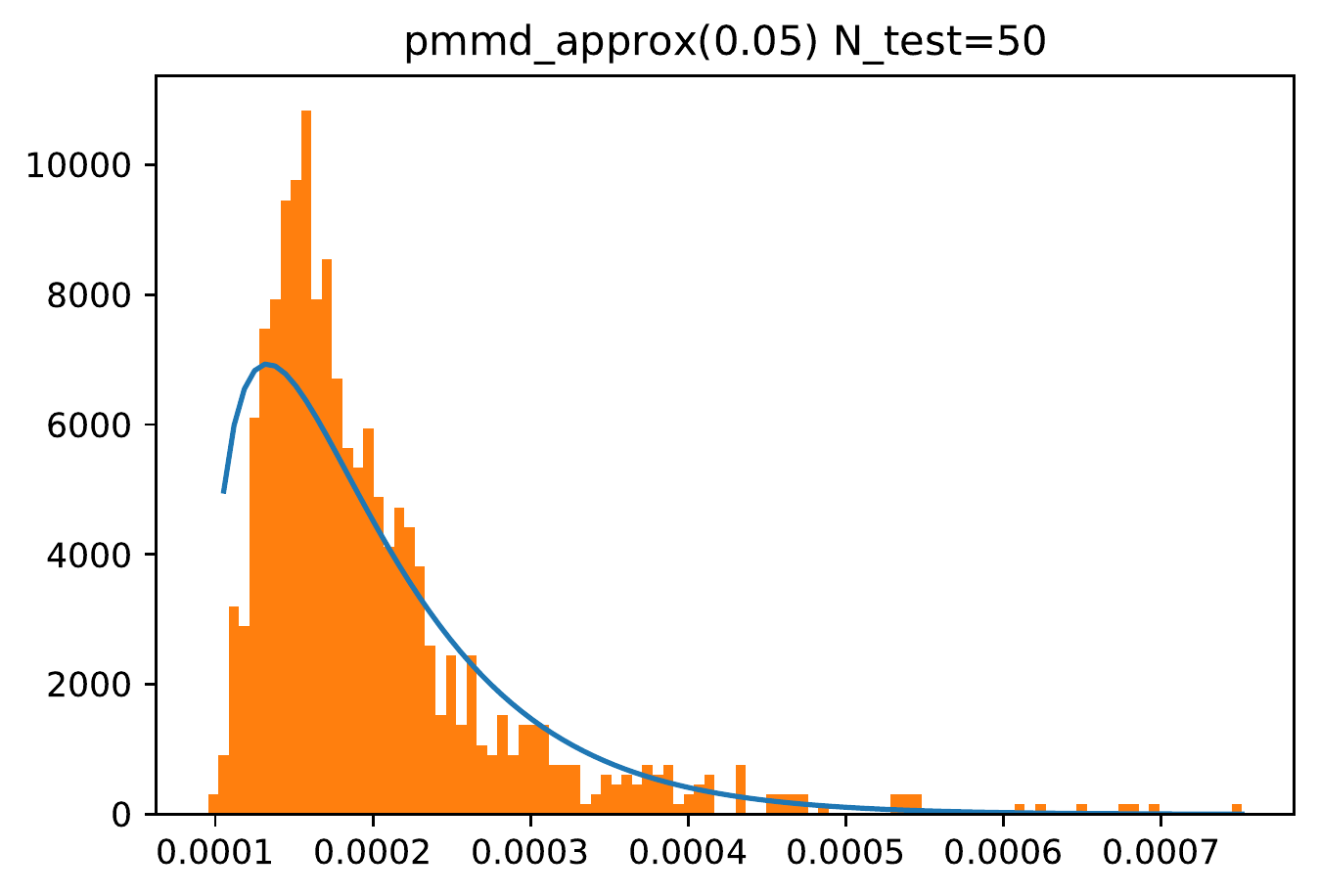}
\includegraphics[width=5cm]{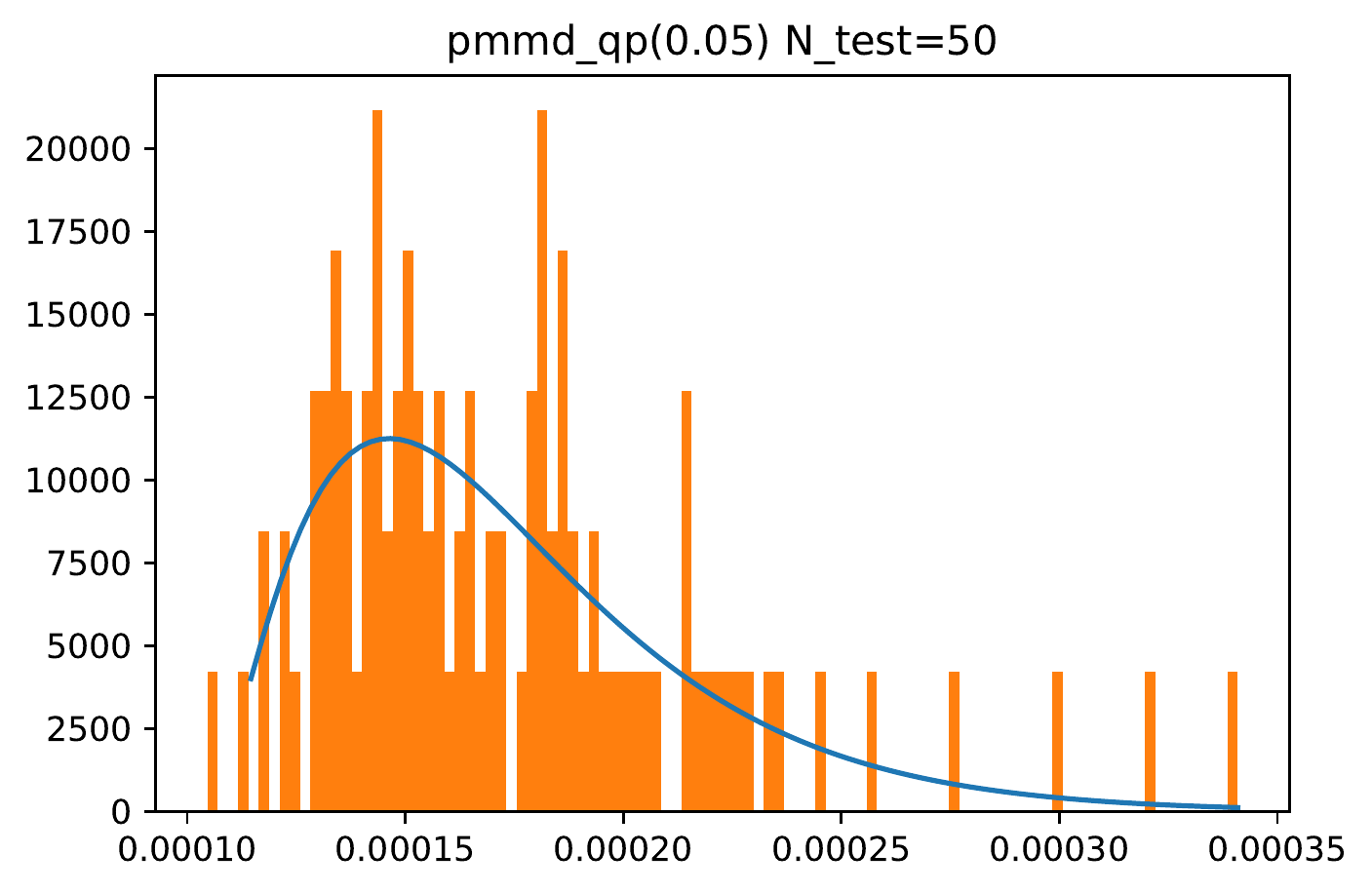}
\caption{\label{fig:goodness-of-fit} Graphical goodness of fit evaluations using the shifted gamma distribution. Note that the QP-based partial MMD is only fit to 100 bootstrap permutations whereas the others use 1000 bootstrap permutations each.}
\end{figure}

Figure \ref{fig:goodness-of-fit} shows the goodness of fit of the fitted shifted gamma distribution as described in Section \ref{sec:bootstrapping}.

One could also fit this instance of the partial Wasserstein distribution with a normal, but the shifted gamma fit seems to give good results in particular at the right tail, which is of interest to us. A possible extension would be to switch between normal approximation and shifted gamma distribution based on the kurtosis.

\begin{figure}
  \includegraphics[width=15cm]{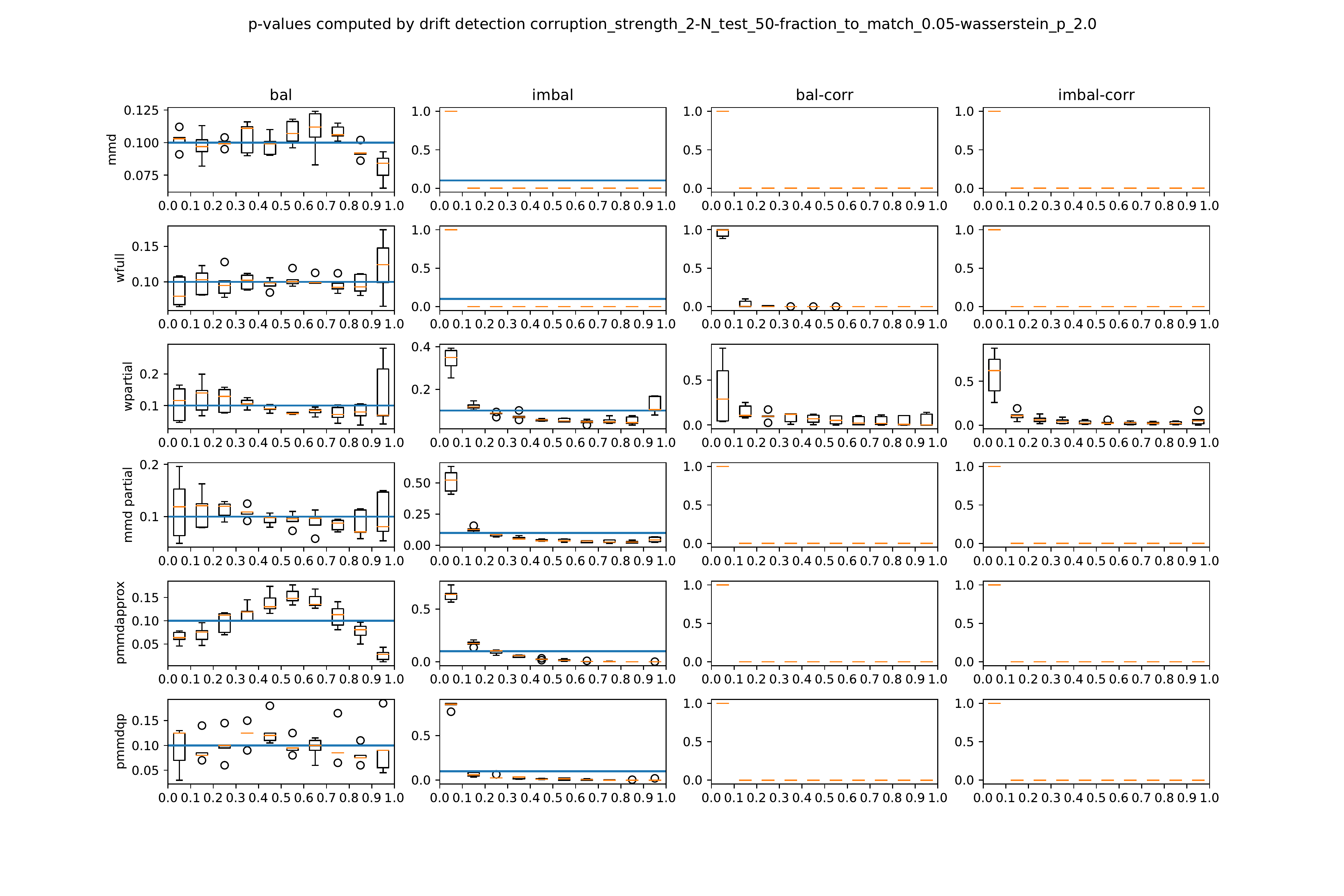}
  \caption{\label{fig:histograms} $p$-value histograms with error bars. Ideally, the left two columns would should a uniform distribution (blue line), while the right two columns would show concentration very much to the left. Here, we do not see the distribution at the lower end, but the ROC curves tell us that for MMD and the full Wasserstein distance, a significant number of draws from the imbalanced value have drift $p$-values lower than those of balanced corrupted samples.}
\end{figure}

We compute the $p$-values for the test statistics to evaluate the mismatch between the botstrapped $p$-values and the actual distributions. For 10 runs of the base experiment, a histogram of them is shown in Figure \ref{fig:histograms}.  We cannot necessarily expect the $p$-value to give the false positive rate in this setting, so we plot histograms of observed $p$-values for corrupted and uncorrupted balanced and imbalanced sample draws, so we regard the $p$-values more as a communication tool. The MMD distance appears to be more robust than the Wasserstein distance. However, as expected, imbalanced data throws off the drift detection by causing very low $p$-values (corresponding to false positives), while the partial Wasserstein distance based drift detection can cope relatively well with these. For the partial Wasserstein distance, corrupted samples can achieve rather high $p$-values, leading to false negatives. This is remedied by using the partial MMD distance in any of the three variants discussed above.

\begin{figure}
\includegraphics[width=5cm]{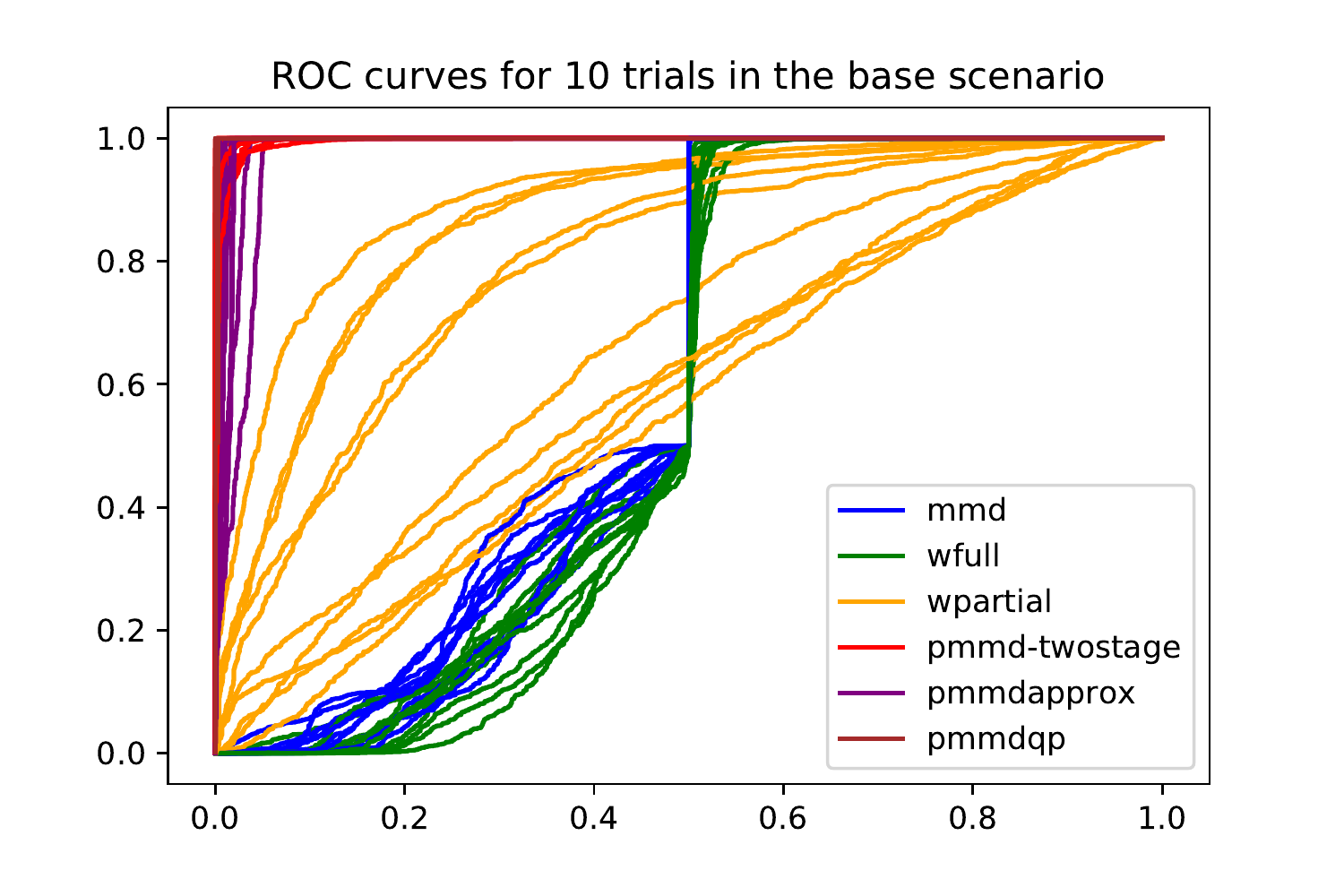}
\includegraphics[width=5cm]{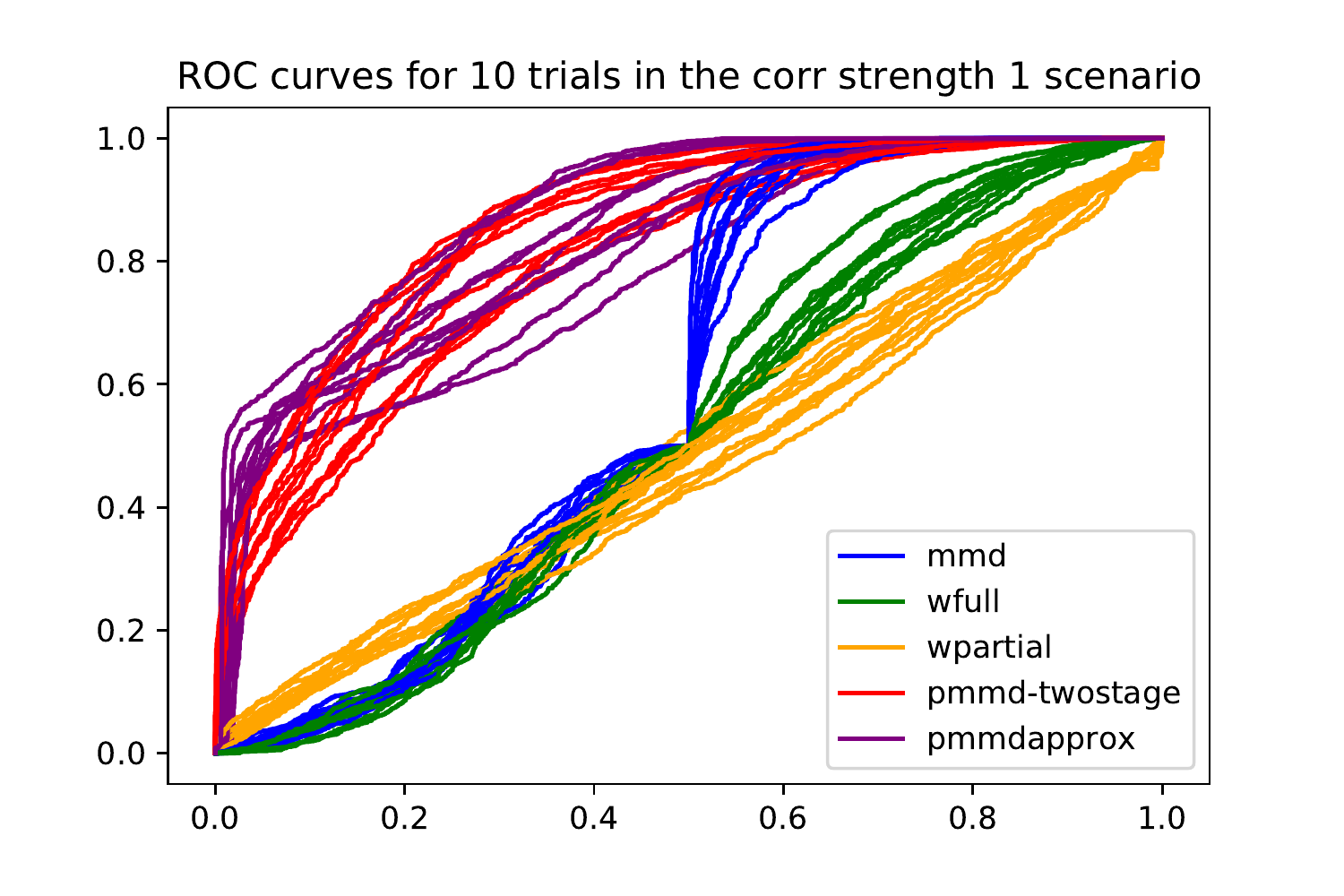}
\caption{\label{fig:roc_curves_base} 10 ROC curves for the detection in the base configuration of the experiment and the low-corruption sensitivity. The full-sample drift detectiors clearly show a large number of false positives at the left, this is from the imbalanced but not corrupted draws. The (partial) Wasserstein distance detector performs much worse than the ones based on the MMD distance, including the two-stage one, and very poorly in the low-corruption experiment. This reflects the spread of the $p$-value mass in the histograms for draws of corrupted samples.}
\end{figure}

For the receiver operating curve (ROC) we pool the balanced and unbalanced examples and then set a $p$-value threshold (which is equivalent to a threshold on the statistic itself). The resut for 10 runs of the base experiments are shown in Figure \ref{fig:roc_curves_base}.
It turns out that the drift detection using the (full sample) Wasserstein distance has more difficulty at detecting some drifts than that based on the MMD distance. However, both exibit many false positives (a very small slope in the beginning), while the partial Wasserstein distance drift detector performs better (in the base case). The advantage of the partial Wasserstein drift detector is more pronounced with stronger severity of the applied corruption but the detection rate is quite bad for low severities. The partial MMD drift detector in any variant performs very well even in the in situations where using the partial Wasserstein distance does not give good drift detectors. This is also reflected in the AUC scores reported below.

\begin{figure}
\includegraphics[width=15cm]{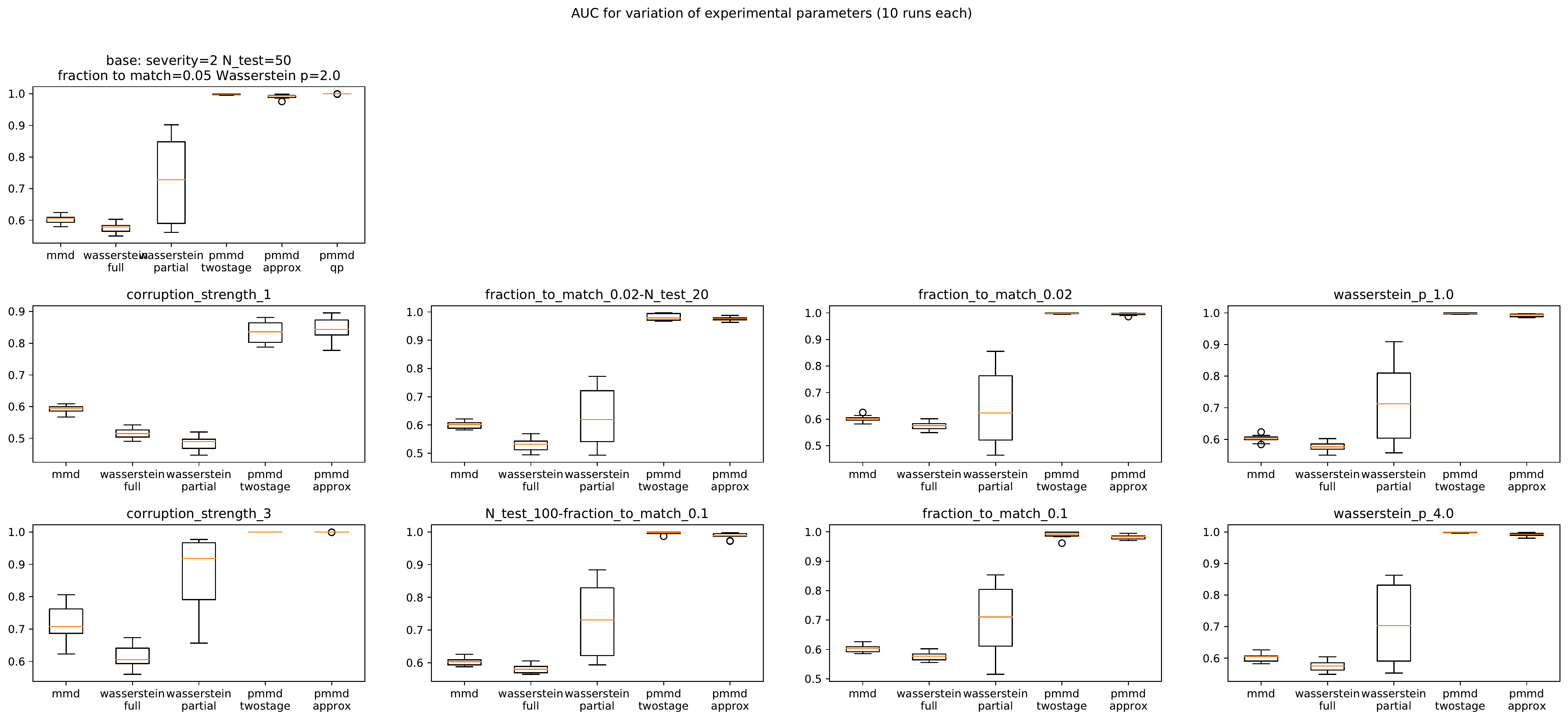}
\caption{\label{fig:auc_sensitivities} Area under the ROC curve for various sensitivities, from 10 runs each. In the leftmost column we vary the severity of the gaussian blur we use to simulate drifted data. In the second column, we change the size of the tested sample and adjust the fraction to match accordingly. In the third column we change the fraction to much but leave the test size at 50 (and reference at 1000), creating a mismatch of sorts. In the rightmost column, we vary the exponent $p$ of the Wasserstein distance definition.}
\end{figure}

To investigate the dependence of our results on experimental parameters, we repeated the experiment while applying the following variations and report the AUC change for 10 runs each. Figure \ref{fig:auc_sensitivities} shows the results.

Reducing the strength of the drift significantly reduces the the detection, strengthening it simplifies the detection of drifted batches. The (partial and full) Wasserstein-based drift detectors' sensitivities perform badly in the base and weak drift scenarios, and only start to work better in the strong drift scenario. The range of drifts in which the MMD-based drift detectors (full and partial) show good results is much broader, including the base scanrio and to some extend also the one with weaker drift.

Running the drift detection on smaller test sets (20 instead of 50) adversely impacts the partial wasserstein drift detector, increasing it (to 100) helps. This is expected. Note that the impact on the MMD-based detector is relatively small, pointing to better robustness and sample efficiency. Interestingly, this is also true for the two-stage partial MMD drift detector which uses the optimal transport coupling from the Wasserstein distance to obtain the weight for the points in the MMD formula.

Introducing a mismatch in the fraction to match forces the partial Wasserstein drift detector to match to several reference samples. It is perhaps surprising that reducing the fraction to match reduces the efficiency, as it would seem that the (partial) coupling could simply be scaled. Increasing the fraction to match appears to improve the detection a bit, but note that we stay at $10\%$, which is the expected fraction of samples of any given class in the reference sample. (As expected, the other detectors are not impacted by a parameter change that does not affect them at all).

Varying the exponent $p$ in the Wasserstein distance form the central $2$ to $1$ appears to not significantly impact the results. In a previous run of the experiments with higher corruption strength, increasing to $4$ appears to to add some reliability for the partial Wasserstein detector, but we can not not observe this here.

As we see in Figure~\ref{fig:auc_sensitivities}, the detection rate of the partial Wasserstein-distance based detector is much more volatile than those of the MMD-based ones. From a more detailed observation, it may be linked to the quality of the feature extractor, caused by the training: If the quality of detection was either poor or good in all 9 sensitivities of a given run with the same feature extractor (which use separate reference distribution drawings and so exclude this as a factor).

All our code is available at \url{https://github.com/torchdrift/partial-wasserstein}.

\section{Attribution}

\begin{figure}
  \includegraphics[width=15cm]{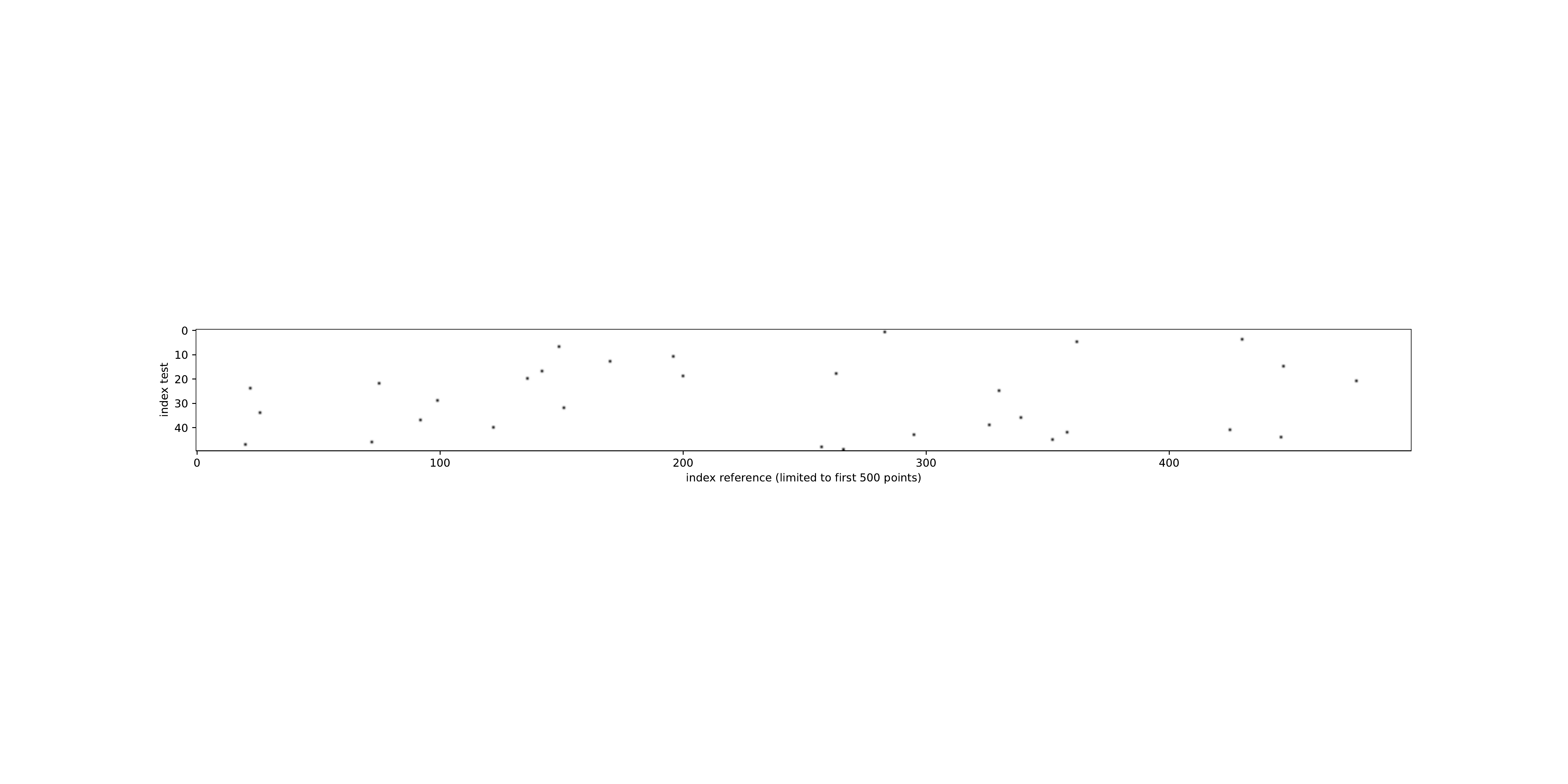}
  \includegraphics[width=15cm]{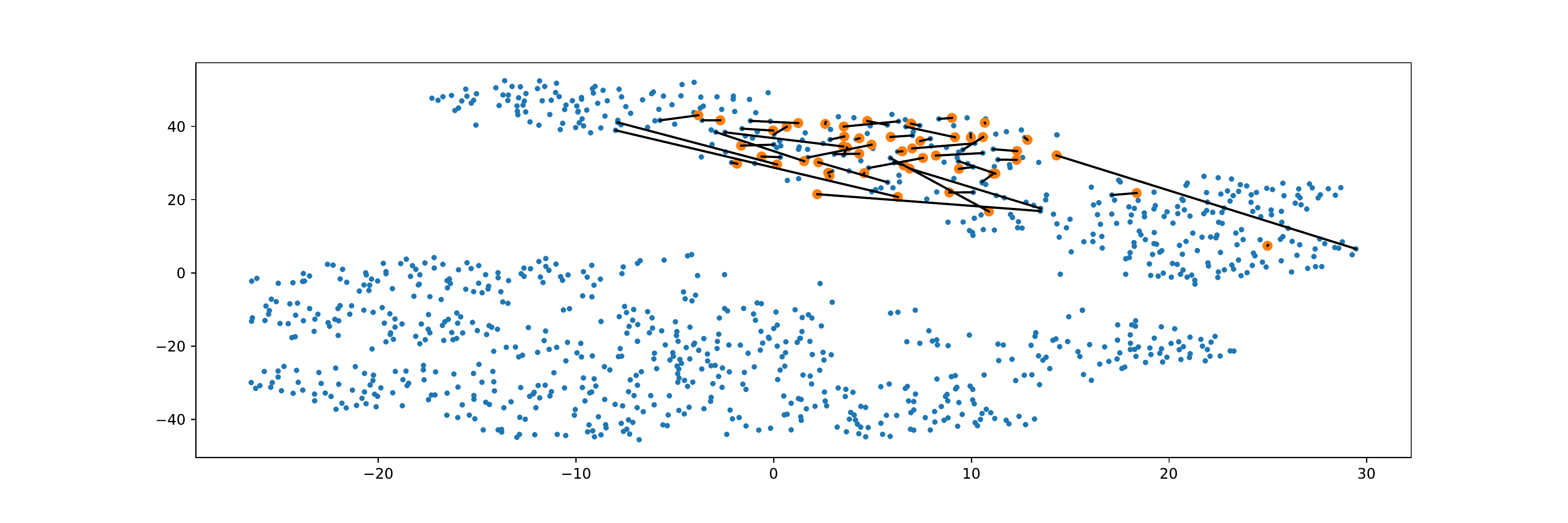}
  
  \caption{\label{fig:matching} The coupling typically is relatively sparsely supported. Top plot: ``heatmap'' of the coupling, the coupling typically has very concentrated mass. (While the optimal coupling is not unique, it is for ``generic'' configurations.)} Bottom plot: Embedding the reference (blue) and an imbalanced test sample (orange) with the coupling.
\end{figure}

For analysis, it is useful to attribute contributions to the drift scrore to each point. For the MMD distance, this can be done using the witness function (see \cite{Gretton}). For the Wasserstein distance, we have even more precise information, as the coupling matches points. Even though a point in the test is mapped to several reference points in general, we expect these to be few or a single one (if the fraction has been calibrated to the relation of test and reference sample sizes). This is more convenient for human inspection (e.g. has the point been mapped to an element of the same class, how large is the distance compared to an average distance etc., which pairings contribute the most to the distance).
Figure \ref{fig:matching} shows a coupling (dots at non-zero weights) as well as a t-SNE projection of the reference and test points along with the matching.

\section{Conclusion}

We have built on the idea to use two-sample testing for detecting data and model drift. We have refined the tested hypothesis from a plain two-sample test to a more task-specific one.
This led us to introduce partial Wasserstein and MMD distance based drift detectiors. By allowing partial matches, we have bridged drift detection to outlier detection.
We have empirically demonstrated its usefulness of our approach in stituations where we cannot expect the tested sample to be exactly representative of the reference distribution.
We have also adjusted the bootstrapping from that directly based on the two-sample-testing to one that is focused on drift-detection, with both theoretical and computational advantages.

The Wasserstein distances and the underlying optimal transport problem allows us to identify a subset of a reference distribution to match against. This is of some use directly, but appears to have a tendency
to not detect weak drifts.
The partial MMD distance and its computationally efficient upper bound approximation through the optimal transport problem underlying the partial Wasserstein distance appears to combine the good sensitivity of the MMD distance for drift detection with the ability to match parts of the reference distribution.

\section{Acknowledgement}

We thank Luca Antiga, Daniele Cortinovis, and Lisa Lozza for inspiring discussions on the subject.

\end{document}